\documentclass[letterpaper, 10 pt, conference]{ieeeconf}  

\IEEEoverridecommandlockouts                              

\usepackage{graphics} 
\usepackage{graphicx}
\usepackage{subcaption}
\usepackage{amsmath} 
\usepackage{amssymb}  

\usepackage{hyperref}
\hypersetup{
	colorlinks=true,
	urlcolor=blue,
}

\title{\LARGE \bf
Trifo-VIO: Robust and Efficient Stereo Visual Inertial Odometry using Points and Lines
}

\author{$^{\dagger}$Feng Zheng, $^{\ast}$Grace Tsai, $^{\dagger}$Zhe Zhang, $^{\ddagger}$Shaoshan Liu, $^{\dagger}$Chen-Chi Chu, and $^{\dagger}$Hongbing Hu
	\thanks{$^{\dagger}$Feng Zheng, Zhe Zhang, Chen-Chi Chu, and Hongbing Hu are with Trifo, Inc., Santa Clara, CA 95054, USA. Email: \texttt{\small \{feng.zheng, zhe.zhang, jason.chu, hongbing.hu\}@trifo.com}\newline
	$^{\ast}$Grace Tsai and $^{\ddagger}$Shaoshan Liu emails are \texttt{\small gstsai@umich.edu}\newline and \texttt{\small shaoshan.liu@perceptin.io} respectively.}%
}

\begin{document}

\maketitle
\thispagestyle{empty}
\pagestyle{empty}

\begin{abstract}

In this paper, we present the Trifo Visual Inertial Odometry (Trifo-VIO), a tightly-coupled filtering-based stereo VIO system using both points and lines.
Line features help improve system robustness in challenging scenarios when point features cannot be reliably detected or tracked, \textit{e.g.} low-texture environment or lighting change.
In addition, we propose a novel lightweight filtering-based loop closing technique to reduce accumulated drift without global bundle adjustment or pose graph optimization.
We formulate loop closure as EKF updates to optimally \textit{relocate} the current sliding window maintained by the filter to past keyframes.
We also present the Trifo Ironsides dataset, a new visual-inertial dataset, featuring high-quality synchronized stereo camera and IMU data from the Ironsides sensor \cite{ironsides} with various motion types and textures and millimeter-accuracy groundtruth.
To validate the performance of the proposed system, we conduct extensive comparison with state-of-the-art approaches (OKVIS, VINS-MONO and S-MSCKF) using both the public EuRoC dataset and the Trifo Ironsides dataset.

\end{abstract}

\section{INTRODUCTION}

Motion tracking is the cornerstone for a wide range of applications, such as robotics, self-driving, AR/VR, \textit{etc}.
Due to complementary properties of cameras and inertial measurement units (IMUs) and the availability of these sensors in smartphones and off-the-shelf plug-and-play devices \cite{ironsides, old_ironsides}, visual-inertial odometry (VIO) has become popular in recent years.
Well-known examples that use VIO are Apple ARKit \cite{ARKit} and Google ARCore \cite{ARCore}.

There are two common ways to categorize VIO approaches.
Based on \textit{when} visual and inertial measurements are fused, VIO approaches can be divided into loosely-coupled and tightly-coupled approaches.
Loosely-coupled approaches \cite{Lynen2013, Ranganathan2007, Weiss2012} estimate motions from images and inertial measurements, independently, and then fuse the two estimates to obtain the final estimate.
Tightly-coupled approaches \cite{imu-preintegration, Leutenegger2015, msckf2.0} fuse visual and inertial data directly at the measurement level to jointly estimate all IMU and camera states.
While loosely coupling is flexible and tends to be more efficient, tightly-coupled approaches generally produce more accurate and robust motion estimates.
Our proposed Trifo-VIO is a tightly-coupled approach.

Based on \textit{how} visual and inertial measurements are fused, VIO approaches can be categorized into filtering-based and optimization-based approaches.
Filtering based approaches \cite{msckf2.0, S-MSCKF} typically employ the Extended Kalman Filter (EKF), where state propagation/prediction is made by integrating IMU measurements, and update/correction is driven by visual measurements.
Contrarily, optimization based approaches \cite{Leutenegger2015, VINS-MONO} use batch nonlinear optimization to directly minimize the errors between integrated motion from IMU measurements and camera motion estimated by the classic reprojection error minimization.
Typically optimization-based approaches are more accurate but computationally more expensive due to repeated linearization.
There are approaches that combines the advantages from both approaches.
For example, PIRVS \cite{pirvs} performs EKF updates iteratively for efficient motion estimation while using optimization (bundle adjustment) at the backend to reduce long-term drifts.
Our proposed Trifo-VIO is an efficient filtering-based VIO, and its accuracy as demonstrated by extensive evaluation is at the same level of or even better than state-of-the-art optimization based approaches.

\begin{figure}[t!]
	\centering
	\includegraphics[width=\columnwidth]{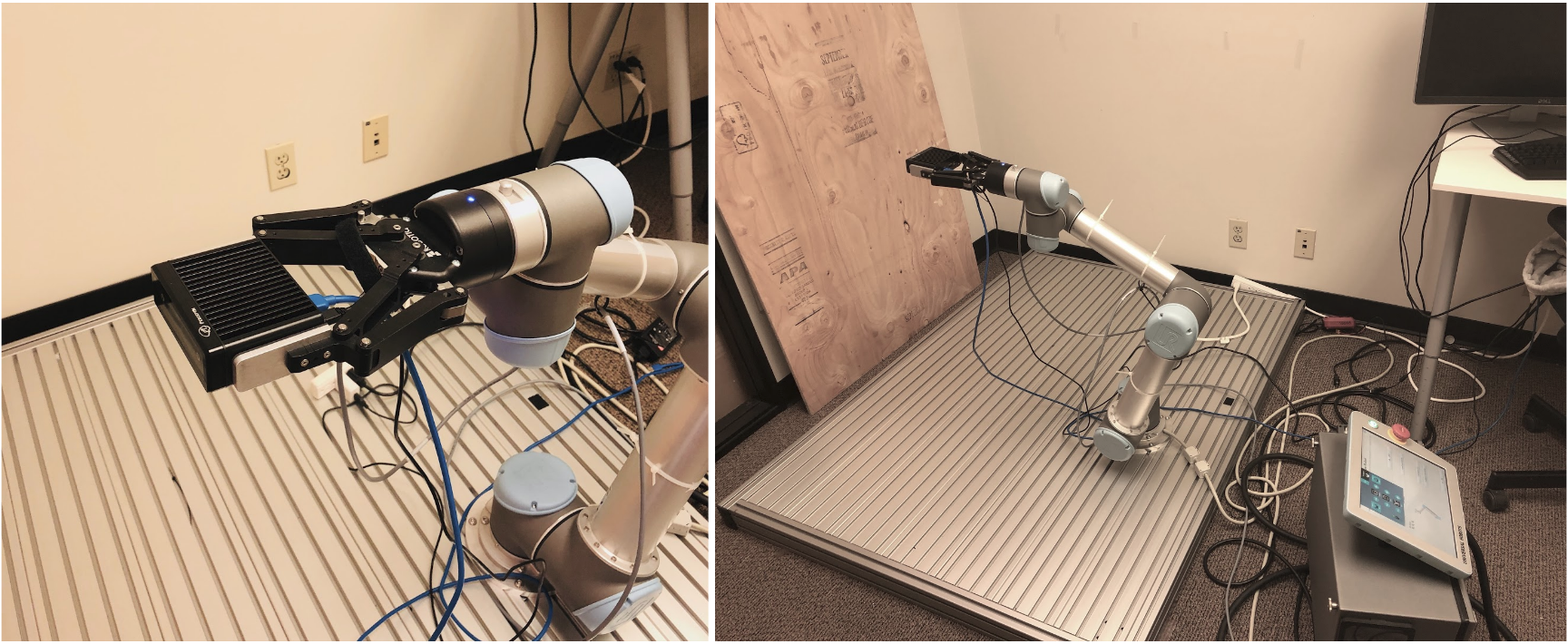}
	\caption{The Trifo Ironsides dataset capture setup.
		The Ironsides sensor \cite{ironsides} outputs synchronized stereo camera and IMU data, at 60Hz and 200Hz respectively.
		The 6-axis robot arm with a working radius of 850 mm provides both motion and millimeter-accuracy groundtruth.
		The dataset contains 9 sequences, featuring various motion types and textures, making it ideal for both evaluation and development.}
	\label{fig:ironsides_capture}
\end{figure}

Most VIO approaches mentioned above only rely on point features, \textit{e.g.} FAST \cite{FAST}, Shi-Tomasi \cite{shi-tomasi}, as intermediate image measurements.
The performance of these approaches suffer considerably in low-texture environments, or in scenarios when point features can not be reliably detected or tracked, \textit{e.g.} lighting change.
Many of such low-texture environments, however, contain planar elements that are rich in linear shapes \cite{Gomez-PL-SLAM}, and the detection of edges is less sensitive to lighting changes in nature.
Therefore, in the proposed Trifo-VIO, in addition to point features, we extract line segment features as useful image measurements to increase the motion constraints available for challenging scenarios, leading to better system robustness.
Both stereo points and line features are processed over a sliding window at cost only linear in the number of features, by using the Multi-State Constraint Kalman Filter (MSCKF) \cite{MSCKF}.

In addition, visual or visual-inertial odometry systems typically operate at faster speed but are more prone to drift compared to SLAM (Simultaneous Localization And Mapping) systems because odometry systems do not maintain a persistent map of the environment.
Therefore, in the proposed Trifo-VIO, we introduce a lightweight loop closing method to reduce long-term drift without any computationally expensive map optimization, \textit{e.g.} bundle adjustment (BA).

We summarize our contributions as follow:
\begin{itemize}
\item To the best of our knowledge, the proposed Trifo-VIO approach is the first tightly-coupled filtering based stereo VIO that uses both point and line features.
\item We introduce a novel lightweight filtering-based loop closure method formulated as EFK updates, which optimally \textit{relocates} the current sliding window maintained by the filter to the detected loops.
\item We conduct extensive evaluation of our Trifo-VIO with comparison to state-of-the-art open-source VIO approaches including OKVIS \cite{Leutenegger2015}, VINS-MONO \cite{VINS-MONO}, and the recent S-MSCKF \cite{S-MSCKF} using both the EuRoC dataset and our new Trifo Ironsides dataset.
\item We release the Trifo Ironsides dataset captured using the Ironsides \cite{ironsides}, a high-quality device with synchronized stereo camera and IMU data, with millimeter-accuracy groundtruth from the robot arm.
The dataset is available at \url{https://github.com/TrifoRobotics/IRONSIDES/wiki/Dataset}.
\end{itemize}

\section{RELATED WORK}

In this section, we review the state of art of odometry or SLAM approaches in terms of line or edge features and loop closure.

PL-SLAM \cite{Pumarola-PL-SLAM} builds on top of ORB-SLAM \cite{Mur-Artal2015} and extend its formulation to handle both point and line correspondences in monocular setup.
In a similar vein, another joint point and line based work \cite{Gomez-PL-SLAM}, termed PL-SLAM as well, aims at stereo camera setting, and additionally introduces a bag-of-words (BoW) place recognition method using both point and line descriptors for loop detection.
In \cite{Tarrio2015}, Tarrio and Pedre propose an edge-based visual odometry for a monocular camera, with simple extension to using rotation prior obtained from gyroscope as regularization term within edge alignment error minimization.
Most recently, Ling \textit{et al.} present a tightly-coupled optimization-based VIO by edge alignment in the distance transform domain \cite{Ling2018}.

Within rich body of filtering-based VIO literatures, there are not many works using edge or line features.
One of the earliest work along this line is \cite{KottasLineVIO}, which uses only line observations to update the filter and also conducts observability analysis.
In \cite{YuLineVIO}, the authors extend \cite{KottasLineVIO} with a new line parameterization which is shown to exhibit better linearity properties and support rolling-shutter cameras.
The edge parametrization introduced in \cite{Yu2017} allows non-straight contours.
Similar to \cite{KottasLineVIO} and \cite{YuLineVIO}, we use straight line segments.

Direct methods, such as LSD-SLAM \cite{Engel2014}, DSO \cite{dso2018}, rely on image intensities at high-gradient regions, which include but are not limited to image region of features and edges.
Usenko \textit{et al.} \cite{Usenko2016} extend the vision-only formulation of LSD-SLAM to tightly couple with IMU by minimizing a combined photometric and inertial energy functional.
ROVIO \cite{rovio-iros2015, rovio-ijrr2017} is a direct filtering-based VIO method, using photometric error of image patches as innovation term in the EKF update.
Our usage of line features, to some extent, lies in between direct and feature-based methods.
Despite the advantage of feature-free operation, direct methods rely on brightness constancy assumption, usually suffering from environment lighting change and camera gain and exposure settings.
In contrast, in particular to ROVIO, we use point reprojection error and point-to-line distance as the filter update innovation instead of photometric error.

Drift is an inhere issue in SLAM and odometry methods.
Loop closure has proven to be effective to correct drift, and state-of-art approaches typically employ global pose graph optimization \cite{Mur-Artal2015, VINS-MONO, Gomez-PL-SLAM}.
In particular, VINS-MONO \cite{VINS-MONO} introduces a two-step loop closure method: (1) local tightly-coupled relocalization which aligns the sliding window with past poses, and (2) global pose graph optimization.
Our lightweight loop closure resembles the first step employed by VINS-MONO, except that we realize it in a filtering framework and we exclude global optimization for efficiency.
To our best knowledge, it is the first tightly-coupled filtering-based loop closure method.
Furthermore, our proposed Trifo-VIO handles both stereo point and line features and loop closure in a consistent filtering framework.

\section{ESTIMATOR DESCRIPTION}

The backbone of our estimator is MSCKF whose key idea is to maintain and update a sliding window of camera poses using feature track observations without including features in the filter state \cite{MSCKF}.
Instead, 3D feature positions are estimated via least-squares multi-view triangulation and subsequently marginalized, which resembles structureless BA to some extent.
The advantage of doing this is considerable reduction of computational cost, making MSCKF's complexity linear in the number of features, instead of cubic like EKF-SLAM \cite{Davison2007}.

We introduce two types of EKF updates: (1) joint point and line features based update to cope with challenging scenarios and to enhance robustness, and (2) loop closing update to reduce accumulated drift.
Filter consistency is ensured by using the right nullspace of the observability Gramian to modify state transition matrix and observation matrix at each propagation and update step, following OC-EKF \cite{OC-VINS}.

\subsection{State Parameterization} 

We follow \cite{MSCKF} and define the evolving IMU state as follows:
\begin{equation}
\textbf{X}_{B} = [ \ _{G}^{B}\textbf{q}^{T} \ \ {\textbf{b}_{g}}^{T} \ \ ^{G}\textbf{v}_{B}^{T} \ \ {\textbf{b}_{a}}^{T} \ \ ^{G}\textbf{p}_{B}^{T} \ \ _{C}^{B}\textbf{q}^{T} \ \ ^{B}\textbf{p}_{C}^{T} \ ]^{T}
\end{equation}
where $_{G}^{B}\textbf{q}$ is the unit quaternion representing the rotation from the global frame $\{G\}$ to the IMU body frame $\{B\}$, $^{G}\textbf{p}_{B}$ and $^{G}\textbf{v}_{B}$ are the IMU position and velocity in the global frame, and $\textbf{b}_{g}$ and $\textbf{b}_{a}$ denote gyroscope and accelerometer biases.
Optionally, we include IMU extrinsics $_{C}^{B}\textbf{q}$ and $^{B}\textbf{p}_{C}$ in the state, which represent the rotation and the translation between the IMU body frame $\{B\}$ and the camera frame $\{C\}$.

At time $k$, the full state of our estimator consists of the current IMU state estimate and $N$ camera poses
\begin{equation}
\hat{\textbf{X}}_{k} = [\ \hat{\textbf{X}}_{B_{k}}^{T} \ \ \hat{\textbf{X}}_{C_{1}}^{T} \ \ ... \ \ \hat{\textbf{X}}_{C_{N}}^{T} \ ]^{T}
\end{equation}
where $\hat{\textbf{X}}_{C} = [_{G}^{C}\hat{\textbf{q}}^{T} \ ^{G}\hat{\textbf{p}}_{C}^{T}]^{T}$ represents the camera pose estimate.

We use the error-state representation in order to minimally parameterize orientation in 3 degrees of freedom (DOF) and to avoid singularities \cite{ESKF}.
Specifically, for the position, velocity, and biases, the standard additive error is employed, while for the orientations, the compositional update $\textbf{q} = \boldsymbol{\delta}\textbf{q} \otimes \hat{\textbf{q}}$ is used, where $\boldsymbol{\delta}\textbf{q}$ is the 3DOF error quaternion as follows
\begin{equation}
\boldsymbol{\delta}\textbf{q} = [ \frac{1}{2} \boldsymbol{\delta} \boldsymbol{\theta} \ 1 ]^{T}
\end{equation}

\subsection{EKF Propagation}

Whenever a new IMU measurement is received, it is used to propagate the EKF state and covariance estimates.
We use the standard continuous-time IMU kinematics model as follows
\begin{align}
_{G}^{B}\dot{\hat{\textbf{q}}}\ &=\ \frac{1}{2} \Omega(\hat{\boldsymbol{\omega}}) _{G}^{B}\hat{\textbf{q}} \\
\dot{\hat{\textbf{b}}}_g\ &=\ \textbf{0}_{3 \times 1} \\ 
^{G}\dot{\hat{\textbf{v}}}\ &=\ R(_{B}^{G}\hat{\textbf{q}})\, \hat{\textbf{a}} + ^{G}\textbf{g} \\
\dot{\hat{\textbf{b}}}_a\ &=\ \textbf{0}_{3 \times 1} \\ 
^{G}\dot{\hat{\textbf{p}}}_{B}\ &=\ ^{G}\hat{\textbf{v}} \\
_{C}^{B}\dot{\hat{\textbf{q}}}\ &=\ \textbf{0}_{3 \times 1} \\
^{B}\dot{\hat{\textbf{p}}}_{C}\ &=\ \textbf{0}_{3 \times 1}
\end{align}
where $\hat{\boldsymbol{\omega}}$ and $\hat{\textbf{a}}$ are angular velocity and linear acceleration from gyroscope and accelerometer respectively with biases removed, $R$ denotes the corresponding rotation matrix of the quaternion, and $\Omega(\hat{\boldsymbol{\omega}}) \in  \mathbb{R}^{4 \times 4}$ is the skew-symmetric matrix formed from the angular rate
\begin{equation}
\Omega(\hat{\boldsymbol{\omega}})\ =\ \begin{bmatrix}
-[\hat{\boldsymbol{\omega}}_\times] & \hat{\boldsymbol{\omega}}\\ 
-\hat{\boldsymbol{\omega}}^{T} & 0 
\end{bmatrix}
\end{equation}
Our discrete-time implementation employs 4th order Runge-Kutta numerical method.
We ignore earth rotation rate in the model as in most MEMS IMUs it cannot be sensed due to gyro bias instability and noise.
For sake of simplicity, we also omit the description of state transition matrix and covariance propagation.
Interested readers please refer to \cite{MSCKF}.

\begin{figure}[t!]
	\centering
	\begin{subfigure}{1.\columnwidth}
		\centering
		\includegraphics[width=\columnwidth]{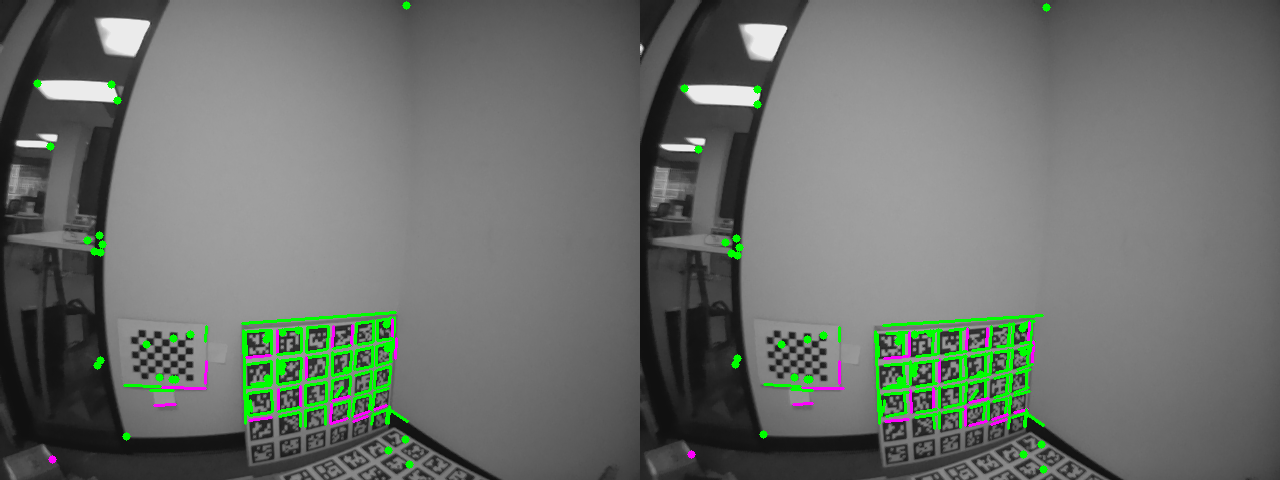}
		\caption{Stereo frame 579}
	\end{subfigure}
	
	\vspace{0.2cm}
	
	\begin{subfigure}{1.\columnwidth}
		\centering
		\includegraphics[width=\columnwidth]{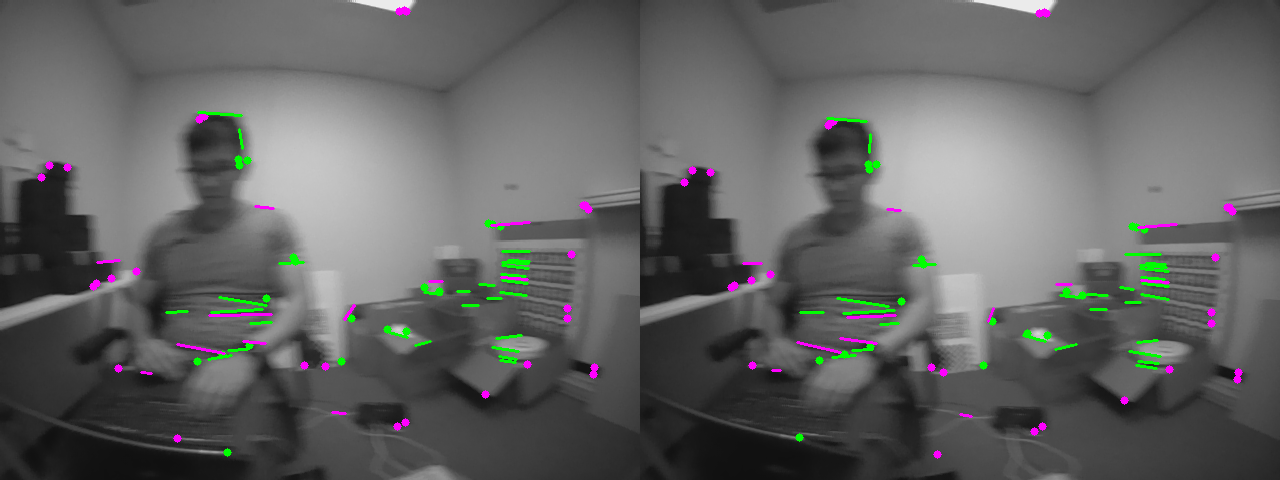}
		\caption{Stereo frame 1262}
	\end{subfigure}
	
	\caption{Stereo point and line features.
		Magenta: new features.
		Green: tracked features.
		Line features help improve system robustness in challenging scenarios, \textit{e.g.} low-texture environment (a) and motion blur (b).
		These two stereo frames are from the Trifo Ironsides dataset PI\_3058.}
	\label{fig:point-line-features}
\end{figure}

\subsection{Measurement Model for Point Features}

In MSCKF, all the \textit{continuous} measurements of the same 3D point, \textit{i.e.} feature tracks, are used to update all involved camera poses that observe the point.
The residual is the standard reprojection error:
\begin{equation}
\textbf{r}_{f_i} = \textbf{z}_{f_i} - \hat{\textbf{z}}_{f_i}
\label{eq:point_rediual}
\end{equation}
where $\textbf{z}_{f_i} = [u_i \ v_i]^T$ is the observation of the \textit{i-th} feature in the image, while $\hat{\textbf{z}}_{f_i}$ is the predicted measurement of the feature from projecting its estimated 3D position $^{G}\hat{\textbf{p}}_{f_i} = [^{G}\hat{X}_{i} \ ^{G}\hat{Y}_{i} \ ^{G}\hat{Z}_{i}]^T$ in global frame into the image based on the estimated camera pose and the projection model as follows
\begin{equation}
\hat{\textbf{z}}_{f_i}\ =\ \pi(\ R(_{G}^{C}\hat{\textrm{\textbf{q}}}) \ (^{G}\textbf{p}_{f_i} - ^{G}\hat{\textrm{\textbf{p}}}_{C}) \ )
\end{equation}
where $\pi$ is the pinhole projection model ($\pi: \mathbb{R}^3 \to \mathbb{R}^2$)
\begin{equation}
\pi(\textbf{p}) = \frac{1}{Z}
\begin{bmatrix}
X \\ 
Y
\end{bmatrix}
\end{equation}
We then linearize Eq. \ref{eq:point_rediual} about the estimates for the camera pose and for the feature position, and calculate the Jacobians with respect to the state and the feature position as $H_{X_{f_i}}$ and $H_{f_i}$ respectively, following \cite{MSCKF}.
After that, we marginalize the feature position via nullspace projection to de-correlate it with the state.

So far, we have described the measurement model for monocular camera.
One tricky part is the estimation of 3D feature positions, which is typically computed by multi-view triangulation in least-squares fashion.
There has to be enough baseline among the cameras observing the same feature in order to do the triangulation.
Therefore, monocular MSCKF cannot estimate the 3D positions of features nor do EKF updates while being static or undergoes rotation dominant motion.
This motivates us to adopt the more practical stereo camera setup to overcome this limitation, from which we can also easily get the true scale.
For stereo feature measurements, we employ a simple yet effective representation, similar to \cite{S-MSCKF},
\begin{equation}
\hat{\textbf{z}}_{f_i}
= 
\begin{bmatrix}
\pi(^{C_1}\hat{\textbf{p}}_{f_i}) \\ 
\pi(^{C_2}\hat{\textbf{p}}_{f_i})
\end{bmatrix}
=
\begin{bmatrix}
\ \pi(\ R(_{G}^{C_1}\hat{\textrm{\textbf{q}}}) (^{G}\textbf{p}_{f_i} - ^{G}\hat{\textrm{\textbf{p}}}_{C_1})\ )\ \\ 
\ \pi(\ R(_{G}^{C_2}\hat{\textrm{\textbf{q}}}) (^{G}\textbf{p}_{f_i} - ^{G}\hat{\textrm{\textbf{p}}}_{C_2})\ )\ 
\end{bmatrix}
\end{equation}
where $\hat{\textbf{z}}_{f_i} \in \mathbb{R}^4$,  $^{C_1}\hat{\textbf{p}}_{f_i}$ and $^{C_2}\hat{\textbf{p}}_{f_i}$ are the estimated 3D positions of the same feature point in left and right camera coordinates respectively, and $\hat{\textbf{X}}_{C_1} = [_{G}^{C_1}\hat{\textbf{q}}^{T} \ ^{G}\hat{\textbf{p}}_{C_1}^{T}]^{T}$ and $\hat{\textbf{X}}_{C_2} = [_{G}^{C_2}\hat{\textbf{q}}^{T} \ ^{G}\hat{\textbf{p}}_{C_2}^{T}]^{T}$ are stereo camera poses at the same timestamp.
Note that the stereo camera is assumed to be calibrated beforehand, and the camera extrinsics relating the left and the right cameras is assumed to be constant.

\subsection{Measurement Model for Line Features}
We now present the measurement model of line features for updating the state estimates.
We denote a line $l_{i}$ in image using point-normal form, $l_{i} = [\textbf{z}_{l_i} \ \vec{\textbf{n}}_{l_i}]$, where $\textbf{z}_i$ is any point on the line and $\vec{\textbf{n}}_{l_i} \in \mathbb{R}^{2\times1}$ is a unit vector denoting line's normal direction in image space.
For a 3D line, $L_{j}$, we over-parameterize it by using two 3D endpoints, $^{G}L_{j} = [^{G}\textbf{p}_b \ ^{G}\textbf{p}_e]$, where $^{G}\textbf{p}_b$ and $^{G}\textbf{p}_e$ are the beginning and ending endpoints on the 3D line in the global frame.

For the line feature residual, $\textbf{r}_{l_i} \in \mathbb{R}^{2\times1}$, we use the point to line distance, as follows
\begin{equation}
\textbf{r}_{l_i} = 
\begin{bmatrix}
(\textbf{z}_{l_i} - \hat{\textbf{z}}_{l_{ib}}) \cdot \vec{\textbf{n}}_{l_i} \\
(\textbf{z}_{l_i} - \hat{\textbf{z}}_{l_{ie}}) \cdot \vec{\textbf{n}}_{l_i}
\end{bmatrix}
\label{eq:line_res_initial}
\end{equation}
where $\hat{\textbf{z}}_{l_{ib}} \in \mathbb{R}^{2\times1}$ and $\hat{\textbf{z}}_{l_{ie}} \in \mathbb{R}^{2\times1}$ are the 2D projections of the beginning and ending endpoints on the 3D line, and $\cdot$ represents dot product.
To conform to the standard form of EKF residual as in Eq. \ref{eq:point_rediual}, we simplify Eq. \ref{eq:line_res_initial} to
\begin{equation}
\textbf{r}_{l_i} = 
\begin{bmatrix}
\vec{\textbf{n}}_{l_i}^T \textbf{z}_{l_i} - \vec{\textbf{n}}_{l_i}^T \hat{\textbf{z}}_{l_{ib}} \\
\vec{\textbf{n}}_{l_i}^T \textbf{z}_{l_i} - \vec{\textbf{n}}_{l_i}^T \hat{\textbf{z}}_{l_{ie}}
\end{bmatrix}
\label{eq:line_res}
\end{equation}
Note $\vec{\textbf{n}}_{l_{ib}}^T \textbf{z}_{l_{ib}}$ produces a scalar number, thus one 3D endpoint results in one dimensional residual.
This is desirable as line features can only provide useful constraints in the normal direction.
Therefore, every 3D line represented by two endpoints produces a two dimensional residual.
The over-parameterization makes sure that, if the projected 3D line and the observed line do not perfectly align, at least in one dimension of the residual it will not be zero.
This holds even when one projected endpoint is accidentally on the observed line.

\begin{figure}[t!]
	\begin{subfigure}[b]{0.48\columnwidth}
		\includegraphics[width=\linewidth]{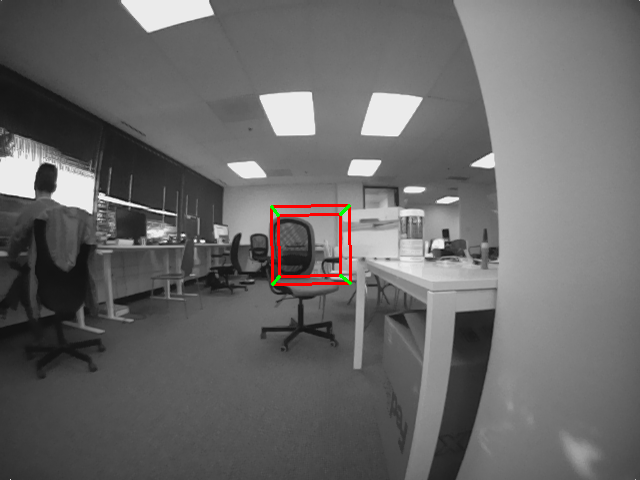}
		\caption{First frame}
		\label{fig:lc-seq-begin}
	\end{subfigure}
	\hfill 
	\begin{subfigure}[b]{0.48\columnwidth}
		\includegraphics[width=\linewidth]{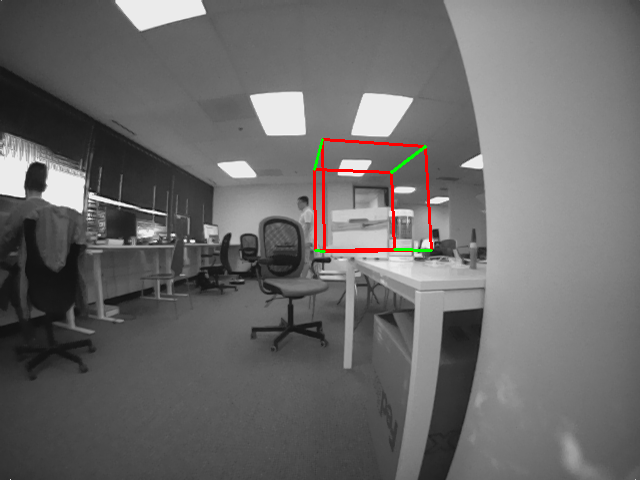}
		\caption{Before loop closure}
		\label{fig:lc-before}
	\end{subfigure}
	
	\begin{subfigure}[b]{0.48\columnwidth}
		\includegraphics[width=\linewidth]{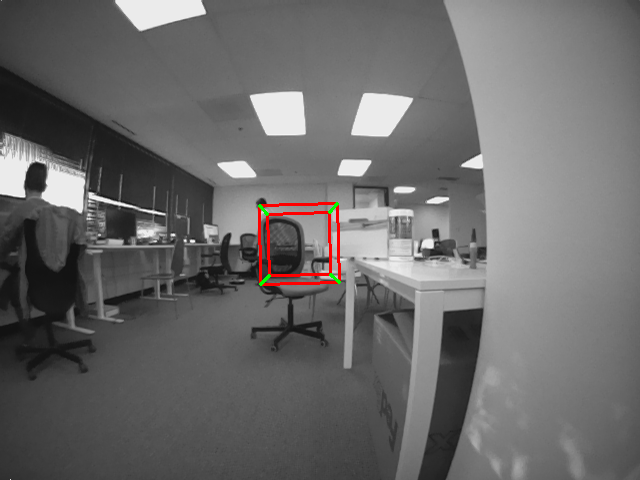}
		\caption{After loop closure}
		\label{fig:lc-after}
	\end{subfigure}
	\hfill 
	\begin{subfigure}[b]{0.48\columnwidth}
		\includegraphics[width=\linewidth]{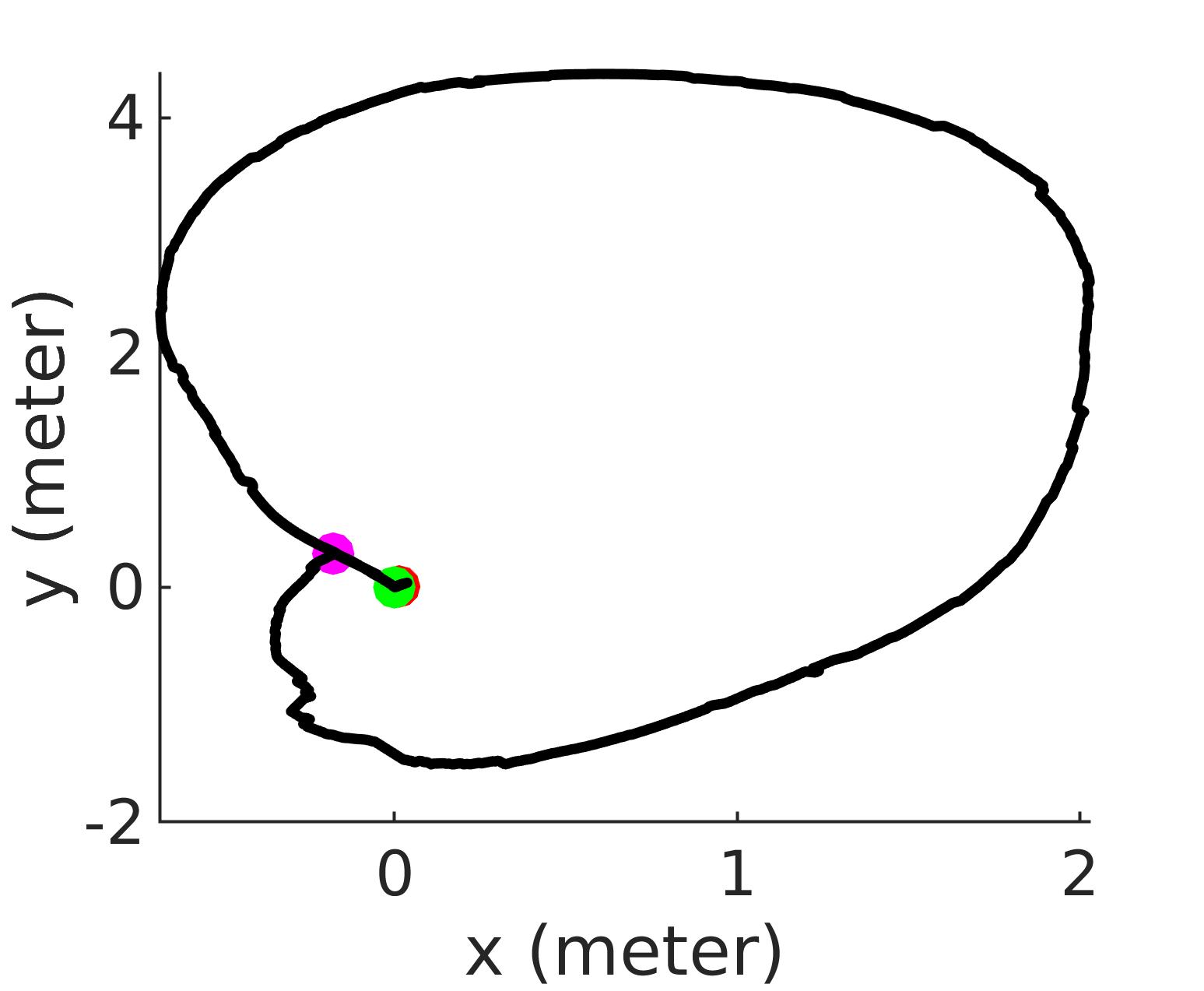}
		\caption{Trajectory}
		\label{fig:lc-traj}
	\end{subfigure}
	
	\caption{Results of loop closing EKF update.
		In this sequence, the Ironsides device is handheld, and we move it around the office and go back to the starting position.
		(a), (b) and (c) are visualizations of device poses at the beginning, around the end before the loop closure and after it, by rendering a fixed-position virtual cube in front of the first camera.
		From the rendered cube, it is evident that the pose drift is effectively corrected.
		In addition, (b) and (c) are not the same frame but a few frames apart, as the drift is corrected progressively rather than immediately.
		(d) shows the trajectory in the horizontal plane, where the red dot indicates the beginning position, and the magenta dot and the green dot are the positions before and after the loop closure, corresponding to (a), (b), and (c) respectively.
		Note that the red dot is largely covered by the green dot, indicating the drift is almost fully corrected.}
	\label{fig:loop_closure}
\end{figure}

Another benefit of this measurement model is that the line feature Jacobian becomes extremely easy to calculate and feature marginalization can be done in the same way as point features.
Under the chain rule, we can derive the Jacobian for each line as follows
\begin{equation}
\textbf{H}_{{l}_i} = 
\begin{bmatrix}
\vec{\textbf{n}}_{l_i}^T \textbf{H}_{l_{ib}} \\
\vec{\textbf{n}}_{l_i}^T \textbf{H}_{l_{ie}}
\end{bmatrix}
\end{equation}
where $\textbf{H}_{l_{ib}} \in \mathbb{R}^{2\times3}$ can be calculated in the same way as point feature Jacobian $H_{f_i}$ and the ``point" here is the beginning endpoint of the line.
Likewise, $\textbf{H}_{l_{ie}} \in \mathbb{R}^{2\times3}$ is the ``point" Jacobian of the line ending endpoint.
Note that $\vec{\textbf{n}}_{l_i}^T \textbf{H}_{l_{ib}} \in \mathbb{R}^{1\times3}$, hence $\textbf{H}_{l_i} \in \mathbb{R}^{2\times3}$.
Similarly, the Jacobian of line feature with respect to the state can be derived as
\begin{equation}
\textbf{H}_{X_{l_i}} = 
\begin{bmatrix}
\vec{\textbf{n}}_{l_i}^T \textbf{H}_{X_{l_{ib}}} \\
\vec{\textbf{n}}_{l_i}^T \textbf{H}_{X_{l_{ie}}}
\end{bmatrix}
\end{equation}
where  $\textbf{H}_{X_{l_{ib}}}$ and $\textbf{H}_{X_{l_{ie}}}$ are the Jacobians of the line's beginning and ending endpoints with respect to the state, and they share the same formula as point features.

To extend the measurement model to the stereo setting is straightforward.
We follow the way we use for stereo point features, and represent the stereo line residual, $\textbf{r}_{l_i} \in \mathbb{R}^4$, as follows
\begin{equation}
\textbf{r}_{l_i} = 
\begin{bmatrix}
\vec{\textbf{n}}_{l_i,1}^T \textbf{z}_{l_i,1} - \vec{\textbf{n}}_{l_i,1}^T \hat{\textbf{z}}_{l_{ib,1}} \\
\vec{\textbf{n}}_{l_i,1}^T \textbf{z}_{l_i,1} - \vec{\textbf{n}}_{l_i,1}^T \hat{\textbf{z}}_{l_{ie,1}} \\
\vec{\textbf{n}}_{l_i,2}^T \textbf{z}_{l_i,2} - \vec{\textbf{n}}_{l_i,2}^T \hat{\textbf{z}}_{l_{ib,2}} \\
\vec{\textbf{n}}_{l_i,2}^T \textbf{z}_{l_i,2} - \vec{\textbf{n}}_{l_i,2}^T \hat{\textbf{z}}_{l_{ie,2}}
\end{bmatrix}
\label{eq:line_res_stereo}
\end{equation}

\subsection{EKF Update: Point and Line Features}

\begin{figure}[t!]
	\centering
	\begin{subfigure}{.495\columnwidth}
		\centering
		\includegraphics[width=\columnwidth]{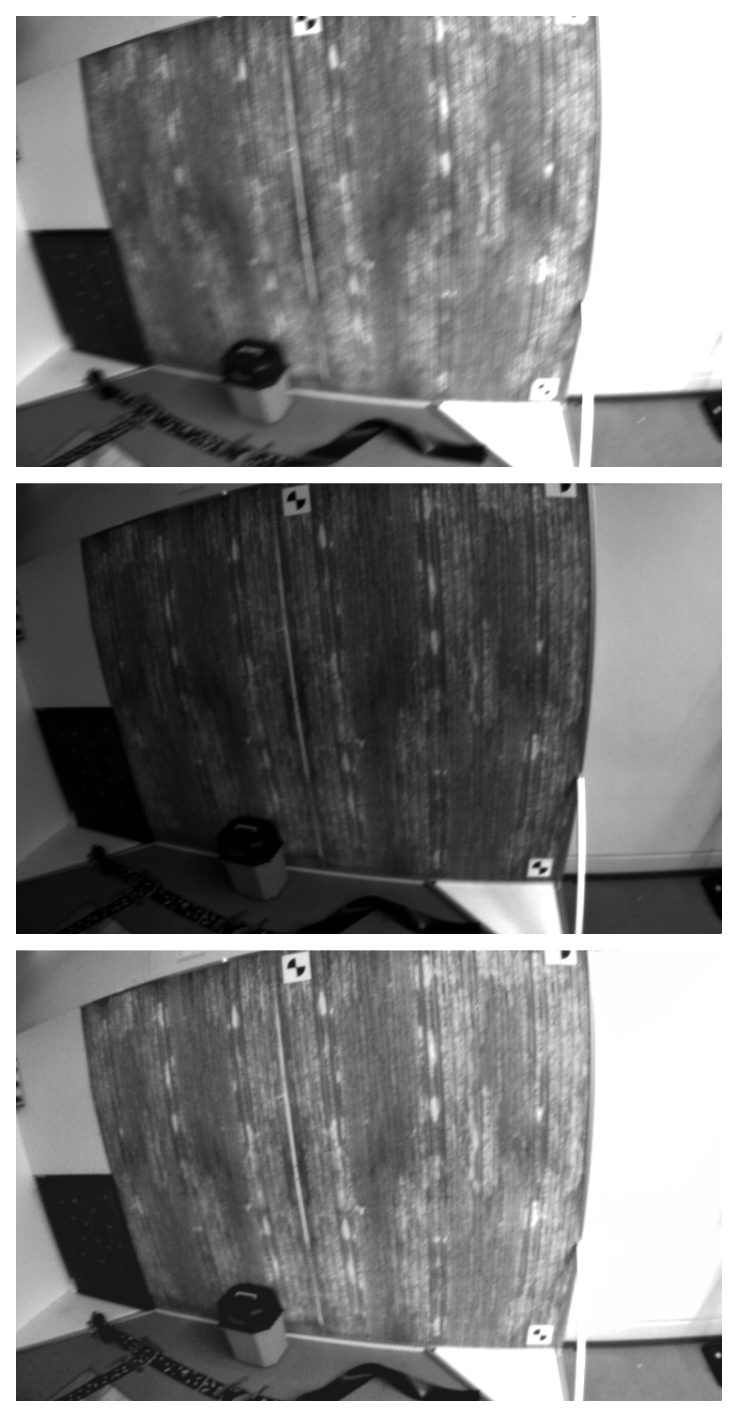}
		\caption{Stereo}
	\end{subfigure}%
	\begin{subfigure}{.495\columnwidth}
		\centering
		\includegraphics[width=\columnwidth]{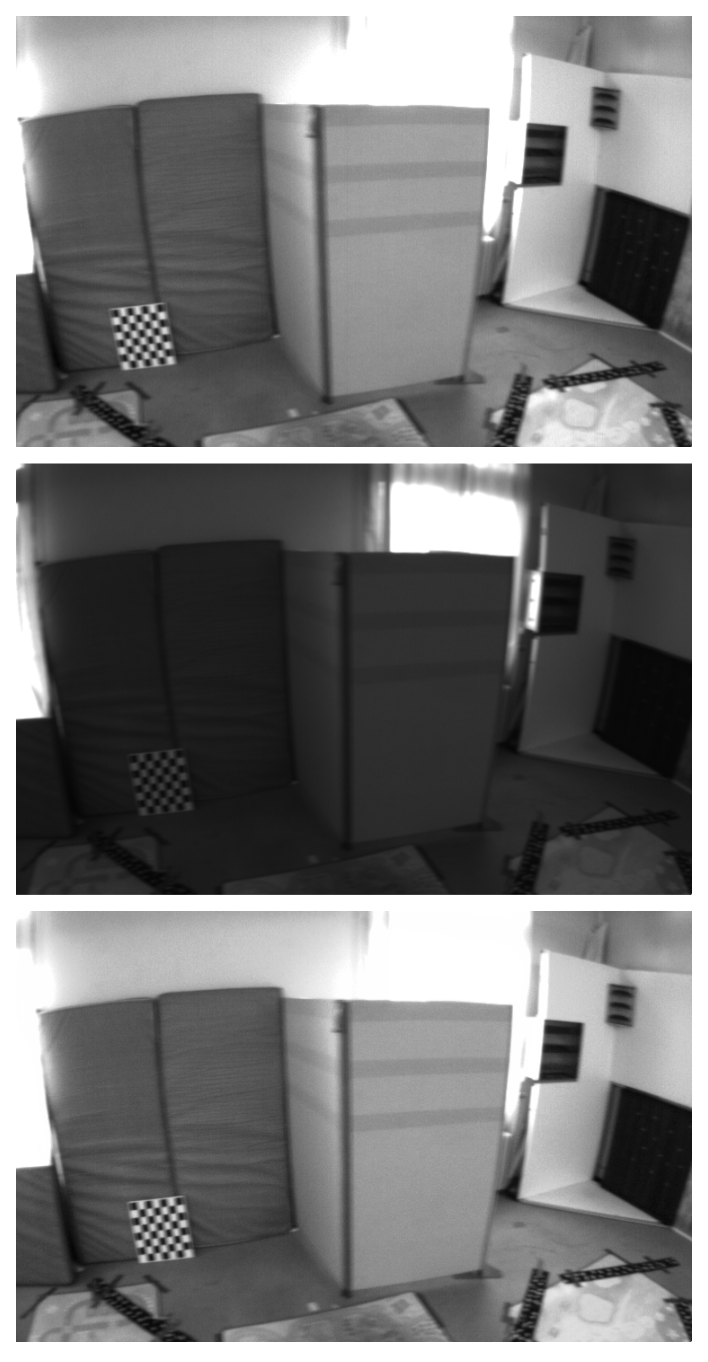}
		\caption{Temporal}
	\end{subfigure}
	\caption{Results of histogram matching (HM) to deal with dramatic brightness mismatch.
		(a) Stereo HM: top - stereo left image, middle - original stereo right image, bottom - stereo right image after HM. 
		The frames are from the EuRoC V2\_03\_difficult dataset.
		(b) Temporal HM: top - left image, middle - original left image at the next timestamp, bottom - left image at the next timestamp after HM.
		The frames are from the EuRoC V1\_03\_difficult dataset.
		It is obvious that middle images in both (a) and (b) exhibit strong brightness difference compared to the top ones, and the HM results at the bottom match the top ones well in terms of overall brightness and distribution.
		Intensity based feature tracking and matching, \textit{e.g.} KLT optical flow \cite{klt-OF}, can significantly benefit from this operation.}
	\label{fig:hist_match}
\end{figure}

We adopt a similar update strategy as \cite{MSCKF}:
whenever a point and/or line feature is no longer tracked, or the sliding window size exceeds the predefined maximum size, EKF update is triggered.
Point and line features are subsequently marginalized since their positions are directly correlated with the state estimate $\hat{X}$.
This makes the algorithm complexity linear in the number of features.
The marginalization is performed by using the left nullspace of feature Jacobian, which cancels out the feature term in the linearized residual.
We then stack the transformed residuals and the state Jacobians of both points and lines to form the final residual and observation matrix.

\subsection{EKF Update: Loop Closure}
\label{sec:loop_close_update}

To reduce accumulated drift while being efficient for resource-constrained platforms which cannot afford global BA, we present a novel lightweight loop closure method, formulated as native EKF updates.
As will be described in Section \ref{sec:loop_detect}, when a new camera state is added to the sliding window, we perform keyframe selection and trigger loop detection in a parallel thread if selected.
If a loop is detected while the keyframe is still in the sliding window, loop closing updates will be triggered.
Otherwise, the keyframe is added to the database along with its feature descriptors and 3D positions.

Since loop detection establishes feature matches between the current keyframe and the past, we use feature positions from the past keyframes for EKF updates instead of re-triangulating them using current poses which suffer from drift.
The update procedure is almost the same as the update with point features, except that we treat 3D positions of such loop closure features as prior knowledge, and thus do not perform feature marginalization.
This makes sense given such ``map'' points have been marginalized in the past along with the keyframes which are inserted into the loop detection database.

As shown in Fig. \ref{fig:loop_closure}, the accumulated drift is effectively corrected by the loop closing update.
A benefit of doing loop closure as EKF updates is that the drift is corrected progressively across multiple consecutive camera frames as long as loop closure features are tracked, rather than immediately which often introduces sudden large jump in the subsequent pose estimate and it is not desirable for closed-loop control, \textit{e.g.} of drones.
While the introduced loop closing update is similar to \cite{Mourikis2009} and \cite{pirvs} where map based update is employed, their maps are either pre-built or online estimated via the costly BA in a separate thread.
To our best knowledge, the introduced tightly-coupled filtering-based loop closure using marginalized ``map'' points is novel.

\section{IMAGE PROCESSING}

In this section, we describe our image processing pipeline for detection and tracking of point and line features.
An example is shown in Fig. \ref{fig:point-line-features}.

For each new image, we track existing point features via KLT optical flow (OF) \cite{klt-OF} and for non-tracked image regions new features are detected via FAST feature detector \cite{FAST}.
We enforce uniform distribution of features in image by spatial binning and maintain a fixed number of high response features in each bin.
To cope with fast motion, we obtain initial guess for optical flow using relative rotation computed from gyroscope measurements.
We use KLT OF for stereo feature matching as well similar to \cite{S-MSCKF} for efficiency.
To reject outliers in both stereo and temporal matching, we use 2-point RANSAC and space-time circular matching \cite{Kitt2010IV}.

For line features, we use Line Segment Detector (LSD) \cite{LSDdetector} to extract line segments.
For each line segment detected, we extract binary descriptor using Line Band Descriptor (LBD) \cite{LBD}.
Both stereo and temporal matching of line features are based on LBD descriptor matching.
To ensure best matches, we perform 4-way consistency check, \textit{i.e.} left to right, right to left, previous to current, and current to previous.
Furthermore, we prune putative matches by checking the length and orientation of lines.

To further enhance the robustness of feature tracking and matching, we introduce a fast brightness check between stereo and temporally consecutive images based on their mean brightnesses, and perform histogram matching to ensure consistent brightness and contrast across stereo and temporal images if necessary.
An example is shown in Fig. \ref{fig:hist_match}.
This considerably boosts stereo feature matching and temporal tracking performance under unfavorable conditions, \textit{e.g.} auto-exposure mismatch between stereo cameras, dramatic lighting change.
This is in contrast to using histogram equalization for each frame as done in \cite{VINS-MONO} and \cite{pirvs}, which incurs more computation and disregards brightness consistency between stereo and temporal images, and may result in over enhancement.

\section{LOOP DETECTION}
\label{sec:loop_detect}

\begin{figure}[t!]
	\centering
	\begin{subfigure}{1.\columnwidth}
		\centering
		\includegraphics[width=\columnwidth]{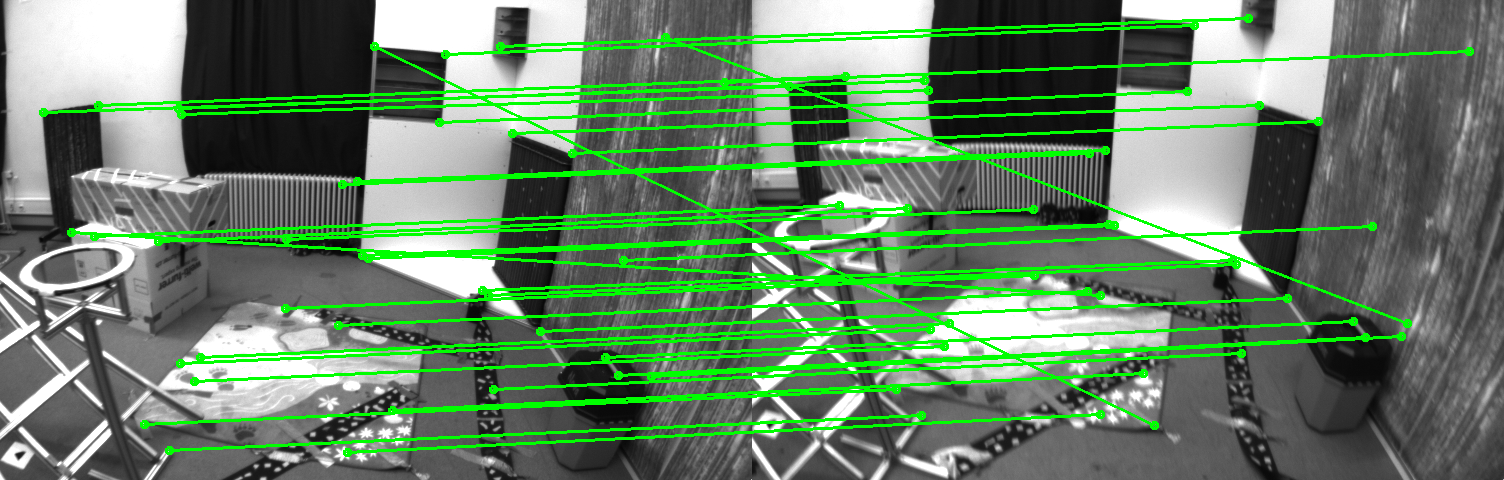}
		\caption{Raw feature matches}
	\end{subfigure}
	
	\vspace{0.2cm}
	
	\begin{subfigure}{1.\columnwidth}
		\centering
		\includegraphics[width=\columnwidth]{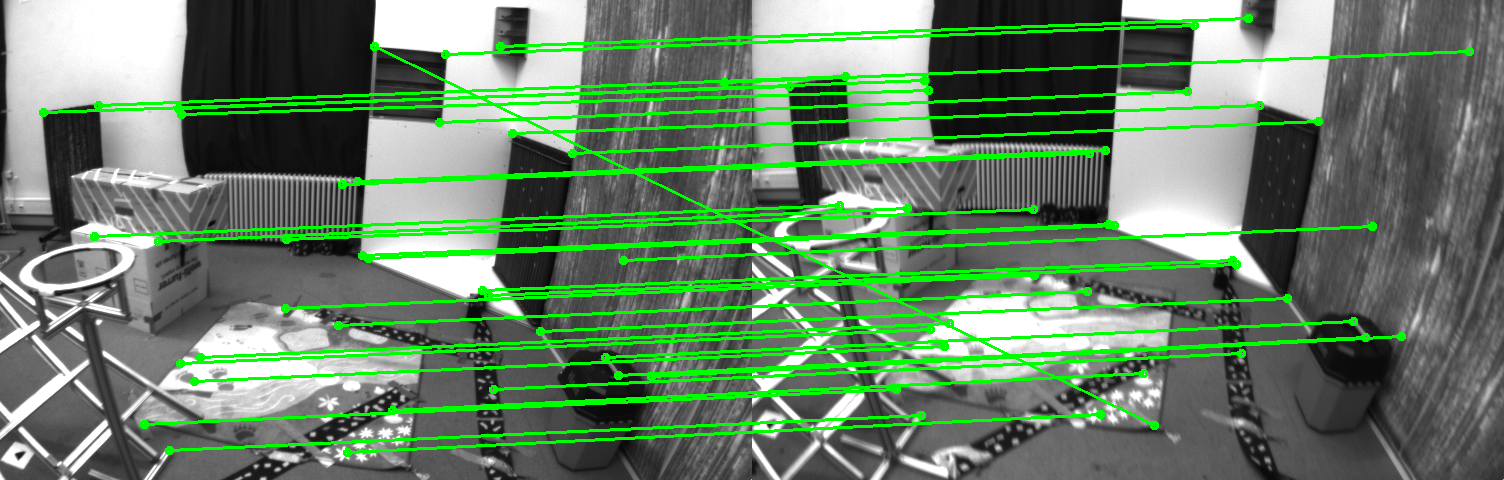}
		\caption{After fundamental matrix RANSAC removal}
	\end{subfigure}
	
	\vspace{0.2cm}
	
	\begin{subfigure}{1.\columnwidth}
		\centering
		\includegraphics[width=\columnwidth]{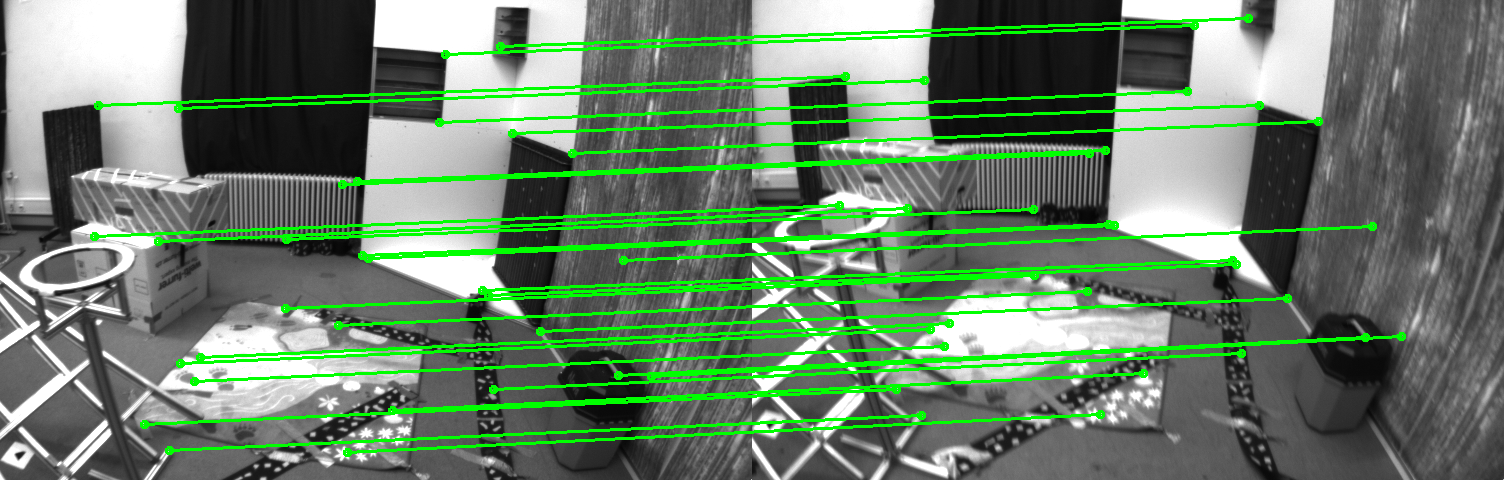}
		\caption{After PnP RANSAC removal}
	\end{subfigure}
	\caption{Two-step outlier rejection for loop-detection feature matches.
		The frames are from the EuRoC V2\_02\_medium dataset.
		(Best viewed in color.)}
	\label{fig:loop_outlier_rejection}
\end{figure}

In this section, we describe our loop detection approach.
For each new image, we do keyframe selection based on the number of features tracked and the pose distance to existing keyframes in the loop detection database.
If a keyframe is selected, we extract ORB descriptors \cite{Rublee2011} for loop detection.
Our loop detection is implemented based on DBoW2 \cite{Galvez-Lopez2012} which is both fast and reliable, and it runs in a parallel thread to the main VIO thread.
For candidate loops, similar to \cite{VINS-MONO}, we perform two-step outlier rejection: 2D-2D fundamental matrix test and 3D-2D PnP test both within the RANSAC framework.
This outlier rejection strategy is effective as shown in Fig. \ref{fig:loop_outlier_rejection}.
If the number of inlier feature matches is above the pre-defined threshold, we mark loop detected and trigger loop closing EKF updates as described in Sec. \ref{sec:loop_close_update}.
If the current keyframe does not contain loops, we add it to the database when it is marginalized from the active sliding window maintained by the filter along with its pose, 2D and 3D positions of features, and their descriptors.
We set a maximum number of keyframes in the database considering memory requirement and detection speed to make sure that it returns result within one camera frame.

\section{EXPERIMENTS}

We conduct two experiments to demonstrate the performance of the proposed Trifo-VIO approach.
Both experiments compare Trifo-VIO to competitive state-of-the-art VIO approaches including OKVIS \cite{Leutenegger2015}, VINS-MONO \cite{VINS-MONO}, and S-MSCKF \cite{S-MSCKF}.
OKVIS and VINS-MONO are optimization based tightly coupled VIO systems, while S-MSCKF is a tightly-coupled filtering-based stereo VIO system closely related to us.
Both OKVIS and S-MSCKF support stereo camera hence we run them in stereo mode, while VINS-MONO is a monocular system.
The first experiment is conducted with the public EuRoC MAV dataset \cite{euroc}, while the second is with our new Trifo Ironsides dataset.
As all comparison approaches contain more or less non-determinism, \textit{e.g.} due to RANSAC, we repeat all experiments five times and report median numbers.

\subsection{EuRoC MAV Dataset}

The EuRoC dataset contains eleven sequences in three categories (MH, V1, V2) collected on-board a Micro Aerial Vehicle (MAV).
We select nine from them so that each category contains 3 datasets.
For comparison approaches, we use their default parameters, as they all have been carefully tuned for the EuRoC dataset.
In addition, we keep the global loop closure on for VINS-MONO, as we want to compare with it in its best form.
Evaluation results are shown in Fig. \ref{fig:euroc_comparison}.
It is evident that Trifo-VIO is among the best performing methods, leading results in MH\_04, MH\_05, V1\_01, and V2\_01.

For the V2\_03 dataset, S-MSCKF produces poor results, mentioned in \cite{S-MSCKF} as well, for the reason that ``the continuous inconsistency in brightness between the stereo images causes failures in stereo feature matching".
Hence, its result is not reported in Fig. \ref{fig:euroc_comparison}.
In contrast, the histogram matching method employed by Trifo-VIO makes it robust to this challenging scenario, as demonstrated in Fig. \ref{fig:hist_match}.
In addition, V2\_03 has about 400 missing frames in the left camera data, resulting in OKVIS tracking failure.
After we prune extra frames from the right camera data, OKVIS runs well.
Note that we use the original V2\_03 dataset for Trifo-VIO evaluation as our approach is robust enough to handle frame drop in either stereo or temporal frames.
VINS-MONO is not affected as it is a monocular approach and uses only left camera data.

\subsection{Trifo Ironsides Dataset}

To further evaluate the performance of Trifo-VIO, we introduce a new public dataset.
The dataset is recorded by the Trifo Ironsides \cite{ironsides} in a robot arm platform as shown in Fig. \ref{fig:ironsides_capture}.
We collect in total 9 sequences, featuring a wide range of motions and environmental conditions, from controlled slow motion around each axis under good visual conditions to fast random motion with motion blur and low texture.
For details, please refer to Table \ref{tab:ironsides_dataset}.
We provide the entire dataset in two formats, ROS bag and zipped format, similar to the EuRoC dataset.
The groundtruth provided by the robot arm is up to millimeter accuracy and precisely synced with the Ironsides sensor.
Therefore, the Trifo Ironsides dataset is ideal for both VIO/SLAM development and evaluation.

\begin{figure}[h!]
	\centering
	\begin{subfigure}{1.\columnwidth}
		\centering
		\includegraphics[width=\columnwidth]{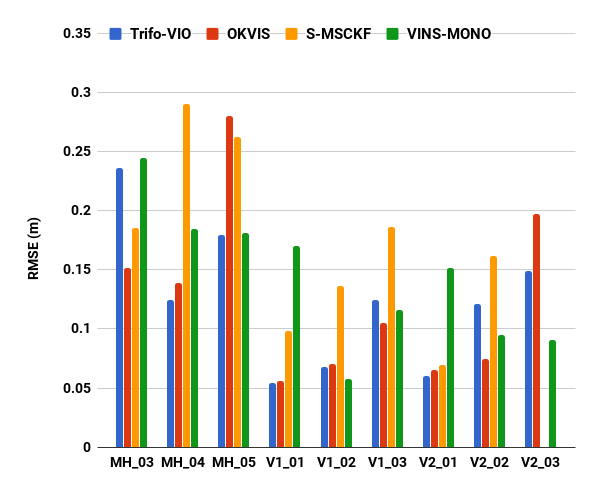}
		\caption{EuRoC dataset}
		\label{fig:euroc_comparison}
	\end{subfigure}
	
	\vspace{0.2cm}
	
	\begin{subfigure}{1.\columnwidth}
		\centering
		\includegraphics[width=\columnwidth]{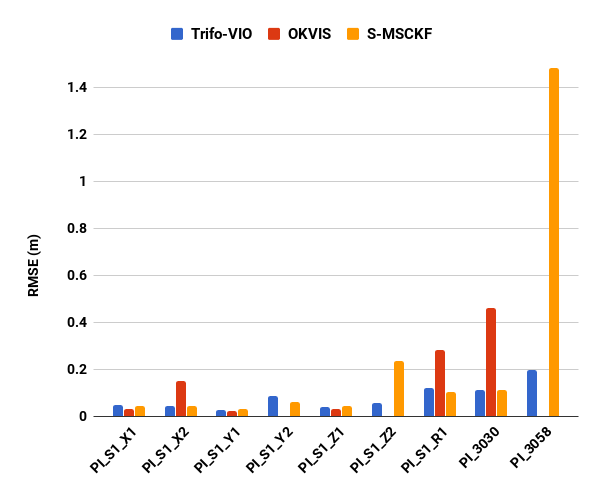}
		\caption{Trifo Ironsides dataset}
		\label{fig:ironsides_comparison}
	\end{subfigure}
	
	\caption{Absolute trajectory RMSE (Root Mean Square Error) results of our Trifo-VIO and competing approaches including OKVIS, S-MSCKF, and VINS-MONO on both the public EuRoC dataset and our new Trifo Ironsides dataset.
		Note that we exclude VINS-MONO from the second evaluation.
		(Best viewed in color.)}
	\label{fig:ate-results}
\end{figure}

The comparison result is shown in Fig. \ref{fig:ironsides_comparison}.
We tune parameters of all approaches to make them perform well as much as possible.
Our Trifo-VIO is consistently among the top two best performing approaches.
S-MSCKF results are close to us, except dataset PI\_3058, where we show significantly better results due to the usage of additional line features.
PI\_3058 is the most challenging sequence in the dataset, containing fast motion and low texture in many parts of the sequence, making it hard for VIO approaches which rely on only point features.
For OKVIS, it performs well for easy sequences (PI\_S1\_X1, PI\_S1\_Y1, and PI\_S1\_Z1) whose dominant motions are slow translation.
However, OKVIS produces poor results for PI\_3058, PI\_S1\_Y2, and PI\_S1\_Z2, hence we omit reporting those numbers.
We notice that OKVIS's feature matching suffers from repetitive textures in the scene.
Note that we exclude VINS-MONO from this comparison as rotation dominant motion at the beginning of many datasets leads to poor initialization and tracking failure.
To improve monocular SLAM initialization, delayed initialization till enough parallax and model selection between fundamental matrix and homography could help \cite{Gauglitz2012, Mur-Artal2015}.

\begin{table}[h!]
	\centering
	\caption{The Trifo Ironsides Dataset}
	\label{tab:ironsides_dataset}
	\begin{tabular}{|c|c|c|c|c|}
		\hline
		\textbf{Dataset} & \textbf{Motion Type} & \textbf{Speed} & \textbf{Texureness} & \textbf{Difficulty} \\ \hline
		PI\_S1\_X1       & Pure X trans. & Slow           & Medium              & Easy                \\ \hline
		PI\_S1\_X2       & Pure X rot.   & Slow           & Medium              & Easy                \\ \hline
		PI\_S1\_Y1       & Pure Y trans. & Slow           & Medium              & Easy                \\ \hline
		PI\_S1\_Y2       & Pure Y rot.   & Slow           & Medium              & Easy                \\ \hline
		PI\_S1\_Z1       & Pure Z trans. & Slow           & Medium              & Easy                \\ \hline
		PI\_S1\_Z2       & Pure Z rot.   & Slow           & Medium              & Easy                \\ \hline
		PI\_S1\_R1       & Random        & Medium         & Medium              & Medium              \\ \hline
		PI\_3030         & Random        & Medium         & Medium              & Medium              \\ \hline
		PI\_3058         & Random        & Fast           & Low                 & Hard                \\ \hline
	\end{tabular}
\end{table}

\section{CONCLUSIONS}

In this work, we have presented Trifo Visual Inertial Odometry (Trifo-VIO), a new tightly-coupled filtering-based stereo visual inertial odometry approach using both point and line features.
Line features help improve system robustness in point-scarce scenarios, \textit{e.g.} low texture and changing light.
Both stereo point and line features are processed over a sliding window at cost only linear in the number of features.
To reduce drift, which is inherent in any odometry approach, we have introduced a novel lightweight loop closure method formulated as native EKF updates to optimally \textit{relocate} the current sliding window maintained by the filter to past keyframes.
All of them (point features, line features, and loop closure) are handled in a consistent filtering-based framework.

We have also presented the Trifo Ironsides dataset, a new public visual-inertial dataset, featuring high-quality synchronized stereo camera and IMU data from the Trifo Ironsides sensor with various motion types and textures and millimeter-accuracy groundtruth.
The extensive evaluation against competitive state-of-the-art approaches using this new dataset and the public EuRoC dataset clearly demonstrate the superior performance of the proposed Trifo-VIO approach.





\section*{ACKNOWLEDGMENT}

Thanks Yen-Cheng Liu for calibrating Ironsides devices and collecting the dataset.
Thanks Chandrahas Jagadish Ramalad for preparing the Trifo Ironsides dataset for release.

\vspace{0.1cm}
\bibliographystyle{IEEEtranS}
\bibliography{refs}

\end{document}